\documentclass{ecai}  % use option [doubleblind] for double blind submission and hiding the authors section
\usepackage{graphicx}
\usepackage{latexsym}
\usepackage{microtype}
\usepackage{algorithmicx,algorithm}
\usepackage{amsfonts}
\usepackage{amssymb}
\usepackage{amsthm,amsmath}
\usepackage{mathrsfs}
\usepackage{makecell}
\usepackage{indentfirst}
\usepackage{multirow}
\usepackage{inconsolata}
\usepackage{multirow}
\usepackage{bbold}
\usepackage{longtable}
\usepackage{graphicx}
\usepackage{arydshln}
\usepackage{booktabs}
\usepackage{setspace}
\usepackage[backref]{hyperref}

\newcommand{\model}{diffusion models with PLMs~}

\begin{document}

\begin{frontmatter}

\title{How Does Diffusion Influence Pretrained Language Models on Out-of-Distribution Data?}

\author[A]{\fnms{Huazheng}~\snm{Wang}}
\author[A]{\fnms{Daixuan}~\snm{Cheng}}
\author[A]{\fnms{Haifeng}~\snm{Sun}\thanks{Corresponding Author. Email: hfsun@bupt.edu.cn.}} % use of \orcid{} is optional \orcid{....-....-....-....}
\author[A]{\fnms{Jingyu}~\snm{Wang}} % use of \orcid{} is optional
\author[A]{\fnms{Qi}~\snm{Qi}} % use of \orcid{} is optional
\author[A]{\fnms{Jianxin}~\snm{Liao}} % use of \orcid{} is optional
\author[A]{\fnms{Jing}~\snm{Wang}\thanks{Corresponding Author. Email: wangjing@bupt.edu.cn.}} % use of \orcid{} is optional
\author[B]{\fnms{Cong}~\snm{Liu}} % use of \orcid{} is optional

\address[A]{State Key Laboratory of Networking and Switching Technology, Beijing University of Posts and Telecommunications}
\address[B]{China Mobile}

\begin{abstract}
Transformer-based pretrained language models (PLMs) have achieved great success in modern NLP. An important advantage of PLMs is good out-of-distribution (OOD) robustness. Recently, diffusion models have attracted a lot of work to apply diffusion to PLMs.
It remains under-explored how diffusion influences PLMs on OOD data.
The core of diffusion models is a forward diffusion process which gradually applies Gaussian noise to inputs, and a reverse denoising process which removes noise. The noised input reconstruction is a fundamental ability of diffusion models.
We directly analyze OOD robustness by measuring the reconstruction loss, including testing the abilities to reconstruct OOD data, and to detect OOD samples.
Experiments are conducted by analyzing different training parameters and data statistical features on eight datasets.
It shows that finetuning PLMs with diffusion degrades the reconstruction ability on OOD data.
The comparison also shows that diffusion models can effectively detect OOD samples, achieving state-of-the-art performance in most of the datasets with an absolute accuracy improvement up to 18\%. 
These results indicate that diffusion reduces OOD robustness of PLMs.
\end{abstract}

\end{frontmatter}

\section{Introduction}
%Natural Language Generation (NLG) is a crucial branch of natural language processing.
Recently, diffusion models~\cite{DBLP:conf/nips/HoJA20} have shown success in vision~\cite{DBLP:journals/corr/abs-2204-06125, DBLP:conf/icml/NicholDRSMMSC22} and text generation~\cite{li2022diffusionlm}.
These models define a forward diffusion process which gradually applies Gaussian noise to real data with a Markov chain.
Then they are trained to learn the reverse denoising process that inverts the forward process by incrementally removing noise from the noised inputs.
Recent works show that transformer-based language models, whether pretrained or finetuned by diffusion, are capable of generating high-quality samples.
This can be achieved by applying noise on discrete token space~\cite{DBLP:journals/corr/abs-2211-15029, DBLP:conf/nips/AustinJHTB21} or continuous latent space~\cite{li2022diffusionlm, DBLP:journals/corr/abs-2211-04236,DBLP:journals/corr/abs-2210-08933}.

Since the training distribution and the test distribution are not usually identical, the out-of-distribution~(OOD) robustness is an important factor for pretrained language models~(PLMs). 
Transformer-based PLMs have shown better robustness than previous language models~\cite{DBLP:conf/acl/HendrycksLWDKS20}. 
It remains under-explored how diffusion influences PLMs. 
For this purpose, we systematically study the OOD robustness from the abilities of reconstructing noised OOD inputs and detecting OOD samples\footnote{Code is available at \url{https://github.com/MaybeLizzy/Diffusion_OOD_Robustness}}.

\textit{Reconstruction Ability.}~The essence of diffusion models is the diffusion and denoising process. Whether the models can reconstruct noised inputs is a fundamental problem.
A direct way to analyze OOD robustness is measuring the reconstruction loss.
We use diffusion to finetune PLMs on in-domain (ID) data following \cite{li2022diffusionlm}, then measure the reconstruction loss on OOD data. If the models are robust, the reconstruction loss should not vary significantly from ID test data to OOD data.
Specifically, diffusion models are sensitive to many issues, such as denoising step, noise ratio, and input size~\cite{DBLP:conf/nips/HoJA20}. We conduct an in-depth analysis from two perspectives, namely training parameters and data statistical features. Experiments show that finetuning PLMs by diffusion leads to poor OOD reconstruction ability.

\textit{OOD Detection.}~In order to investigate to which extent diffusion influences PLMs on OOD data,
we use \model to do OOD sample detection and perform a numerical comparison, by proposing a diffusion-based OOD detection method.
It reconstructs noised inputs and measures the average token reconstruction loss with a threshold to detect OOD samples. 
%Input is regarded as ID data if loss is less than the threshold, otherwise it is regarded as OOD data. 
Comparing with different widely-used baselines, \model achieve state-of-the-art performance on most datasets with an absolute accuracy improvement up to 18\%. 
The improvement on OOD detection, on the other hand, demonstrates the reduction in the robustness of \model.

To summarize, our work reveals that finetuning PLMs with diffusion reduces the OOD robustness. The extent of the impact is related to factors including training step, model size, data diversity, and sentence length (\S~\ref{sec:ability}). We also provide adapted solutions to improve the robustness. Based on the robustness degradation, we propose a diffusion-based OOD detectors which provides high-accuracy detections by simply computing the reconstruction loss (\S~\ref{sec:detection}).

\section{How We Test Reconstruction Ability}

\subsection{Datasets}
Following~\cite{DBLP:conf/emnlp/Zhou0C21}, we use four classification-task datasets as ID data, including  SST2~\cite{DBLP:conf/emnlp/SocherPWCMNP13}, IMDB~\cite{DBLP:conf/acl/MaasDPHNP11}, 20NG~\cite{DBLP:conf/icml/Lang95} and TREC-10~\cite{DBLP:conf/coling/LiR02}.
Any pair of the above datasets can be regarded as
OOD to each other, except for IMDB and SST2 which belong to the same task category.
Besides, we also select four additional datasets as the OOD datasets, including RTE~\cite{DBLP:conf/mlcw/DaganGM05}, MNLI~\cite{DBLP:conf/naacl/WilliamsNB18}, WMT16~\cite{DBLP:conf/wmt/BojarCFGHHJKLMN16} and Multi30K~\cite{elliott2016multi30k}.
We take the test splits in those OOD datasets for testing and report the average sequence lengths of the test splits. The statistics of the datasets are shown in Table~\ref{dataset}. %(More detailed information refers to \hyperref[a2]{A.2}.)

\subsection{Methods}
A diffusion model~\cite{DBLP:conf/nips/HoJA20} is a latent variable model that builds a transformation from Gaussian distribution to data distribution through a multi-step denoising process. 
Given a data distribution $\boldsymbol{x}_0 \sim q(\boldsymbol{x}_0)$, at timestep $t$, $\boldsymbol{x}_t$ is produced by a diffusion process which satisfies a Markov chain according to a variance schedule $\beta_1, \beta_2,...,\beta_T$ $\in(0,1)$,
\begin{equation}
\setlength{\abovedisplayskip}{3pt}
\setlength{\belowdisplayskip}{3pt}
q(\boldsymbol{x}_{1:T}|\boldsymbol{x}_0) := \prod \limits_{t=1}^T q(\boldsymbol{x}_t|\boldsymbol{x}_{t-1}),
\label{xt}
\end{equation}
\begin{equation}
\setlength{\abovedisplayskip}{3pt}
\setlength{\belowdisplayskip}{3pt}
q(\boldsymbol{x}_t|\boldsymbol{x}_0) = N(\boldsymbol{x}_t; \sqrt{\overline{\alpha}_t}\boldsymbol{x}_0, (1-\overline{\alpha}_t)\boldsymbol{I}),
\label{xt}
\end{equation}
where $0 \leq t \leq T$, $\alpha_t := 1 - \beta_t$, $\overline{\alpha}_t := \prod_{s=1}^t \alpha_s$, $T$ is the max iteration step.
A closed form of $\boldsymbol{x}_t$ can be calculated for any arbitrary $t \geq 1$
\begin{equation}
\setlength{\abovedisplayskip}{3pt}
\setlength{\belowdisplayskip}{3pt}
\boldsymbol{x}_t = %\sqrt{\alpha_t}\boldsymbol{x}_{t-1} + \sqrt{1-\alpha_t}\epsilon_t = 
\sqrt{\overline{\alpha}_t}\boldsymbol{x}_0 + \sqrt{1-\overline{\alpha_t}}\epsilon,
\label{xt}
\end{equation}
where $\epsilon\in N(0,1)$.
When the data is fully noised, $\boldsymbol{x}_T$ is close to an isotropic Gaussian, $\boldsymbol{x}_T \sim N(0,1)$.
Note that the predefined forward process $q$ contains no trainable parameters. 

In the denoising process, given $\boldsymbol{x}_t$ and $t$, the model is trained to reverse the diffusion process iteratively and to reconstruct data.
Each denoising transition $\boldsymbol{x}_t \rightarrow \boldsymbol{x}_{t-1}$ is parametrized by
\begin{equation}
\setlength{\abovedisplayskip}{3pt}
\setlength{\belowdisplayskip}{3pt}
p_\theta(\boldsymbol{x}_{t-1}|\boldsymbol{x}_t) = N(\boldsymbol{x}_{t-1} | \boldsymbol{\mu}_\theta(\boldsymbol{x}_t, t), \boldsymbol{\Sigma}_{\theta}(\boldsymbol{x}_t,t)),
\label{xt}
\end{equation}
where $\boldsymbol{\mu}_\theta$ and $\boldsymbol{\Sigma}_{\theta}$ are predicted by a transformer-based PLMs.
Following~\cite{li2022diffusionlm}, instead of predicting the mean posterior $\mu_\theta(\boldsymbol{x}_t,t)$, we train a neural network to predict $\boldsymbol{x}_0$ in every term and simplify the loss to a sum of mean-squared errors between the ground truth data $\boldsymbol{x}_0$ and its estimates $\hat{\boldsymbol{x}_0}$
\begin{equation}
\setlength{\abovedisplayskip}{3pt}
\setlength{\belowdisplayskip}{3pt}
L_{d}(\theta) = \mathbb{E}_{t,\boldsymbol{x}_0}\left[\Vert \boldsymbol{x}_0 - \hat{\boldsymbol{x}_0}(\boldsymbol{x}_t,t,\theta) \Vert^2 \right].
\label{loss2}
\end{equation}

To apply diffusion on transformer-based PLMs, we define the diffusion process on the continuous embedding space.
Considering a sequence $z = \{w_1, w_2,..., w_n\}$, where each token $w_i \in \mathbb{R}^v$ has an associated embedding $e_i \in \mathbb{R}^d$, the discrete-to-continuous step is then defined as $q_\phi(\boldsymbol{x}_0|\boldsymbol{w})=N(\boldsymbol{E}\boldsymbol{w}, \sigma_0^2 \boldsymbol{I})$, where $d$ is the dimension of input space, $\boldsymbol{E}$ is a matrix of all embeddings and $\sigma_0$ is a constant scale factor with a similar order of magnitude as $\beta_1$.
To map the predicted vector back to tokens, we define a reverse continuous-to-discrete step $p_\theta(\boldsymbol{w}|\boldsymbol{x}_0)=\prod_{i=1}^n p_\theta(\boldsymbol{w}_i|\boldsymbol{x}_i)$, where $p_\theta(\boldsymbol{w}_i|\boldsymbol{x}_i)$ is the softmax probability of token $i$ with logit $\boldsymbol{x}_i$.
A simple cross-entropy loss is then added to maximise $p_\theta(\boldsymbol{w}|\boldsymbol{x}_0)$
\begin{equation}
\setlength{\abovedisplayskip}{3pt}
\setlength{\belowdisplayskip}{3pt}
L_{c}(\theta) = \mathbb{E}_{\boldsymbol{w},\boldsymbol{x}_0}\left[-log p_\theta(\boldsymbol{w}|\boldsymbol{x}_0) \right].
\label{loss3}
\end{equation}
The reconstruction loss function is then defined by
\begin{equation}
\setlength{\abovedisplayskip}{3pt}
\setlength{\belowdisplayskip}{3pt}
L_{recon} = L_{d} + L_{c}.
\label{loss3}
\end{equation}

 \begin{table}
\centering
\small
%\small
\begin{tabular}{c|cccc}	%\hline
\hline
Dataset & Train & Dev & Test  & Avg len\\
\hline
SST2 & 67349 & 872 &  1821 &  25\\
IMDB & 22500 & 2500 & 25000 & 295 \\
TREC-10 & 4907 & 545 & 500 & 10\\
20NG & 15056 & 1876 & 1896 &  865\\
\hline
MNLI & - & - & 19643 & 38 \\
RTE & - & - & 3000 & 65  \\
WMT16 & - & - & 2999 & 25 \\
Multi30K & - & - & 2532 & 15 \\
\hline
\end{tabular}
\caption{Dataset Statistics.}
\label{dataset}	
\end{table}

\subsection{Experimental Setups}
We train the diffusion model on the training splits of the four aforementioned ID datasets respectively and use validation splits for evaluation to avoid overfitting.
When it comes to inference, the other test splits of the remaining datasets are treated as OOD data.
We implement the model upon pretrained RoBERTa-Large~\cite{DBLP:conf/emnlp/WolfDSCDMCRLFDS20}. 
The learning rate is set to be $5e-5$ with a linear decay rate towards 0.
We set $\beta$ to constants as~\cite{li2022diffusionlm} and~\cite{DBLP:conf/nips/HoJA20} did. It increases linearly from $1e-4$ to $0.02$, standing for the amount of noise added at diffusion step $t$.
And the batch size is 16.
Each model is trained for $s=80k$ steps, optimized with Adam~\cite{DBLP:journals/corr/KingmaB14}.
$T$ is set to be 1000.
Reconstruction loss is calculated as the average loss of each word in the whole sentence.

\section{Reconstruction ability of Diffusion Models with PLMs}
\label{sec:ability}

\subsection{Impact of Model Hyper-parameters and Structures}
\label{sec:hyper}

\textbf{OOD data is more sensitive to higher noise level.}
In the process of diffusion, the added noise level, i.e. the diffusion step $t$, is pivotal in restoring noised inputs.
%Normally, it is harder for models to reconstruct the original sentence when $t$ is getting larger~\cite{DBLP:journals/corr/abs-2211-07740}.
However, under which $t$ the model starts to produce dissimilar outputs of ID and OOD data has not yet been studied.
Hence, in this experiment, we explore how reconstruction loss varies with the increase of $t$.
%However, which $t$ determines the greatest difference of reconstruction ability on ID/OOD data is still under explored.
%Hence, we explore how reconstruction loss varies with the increase of $t$ (Fig.\ref{t}).
\begin{figure}
\centering
\includegraphics[width=0.425\textwidth]{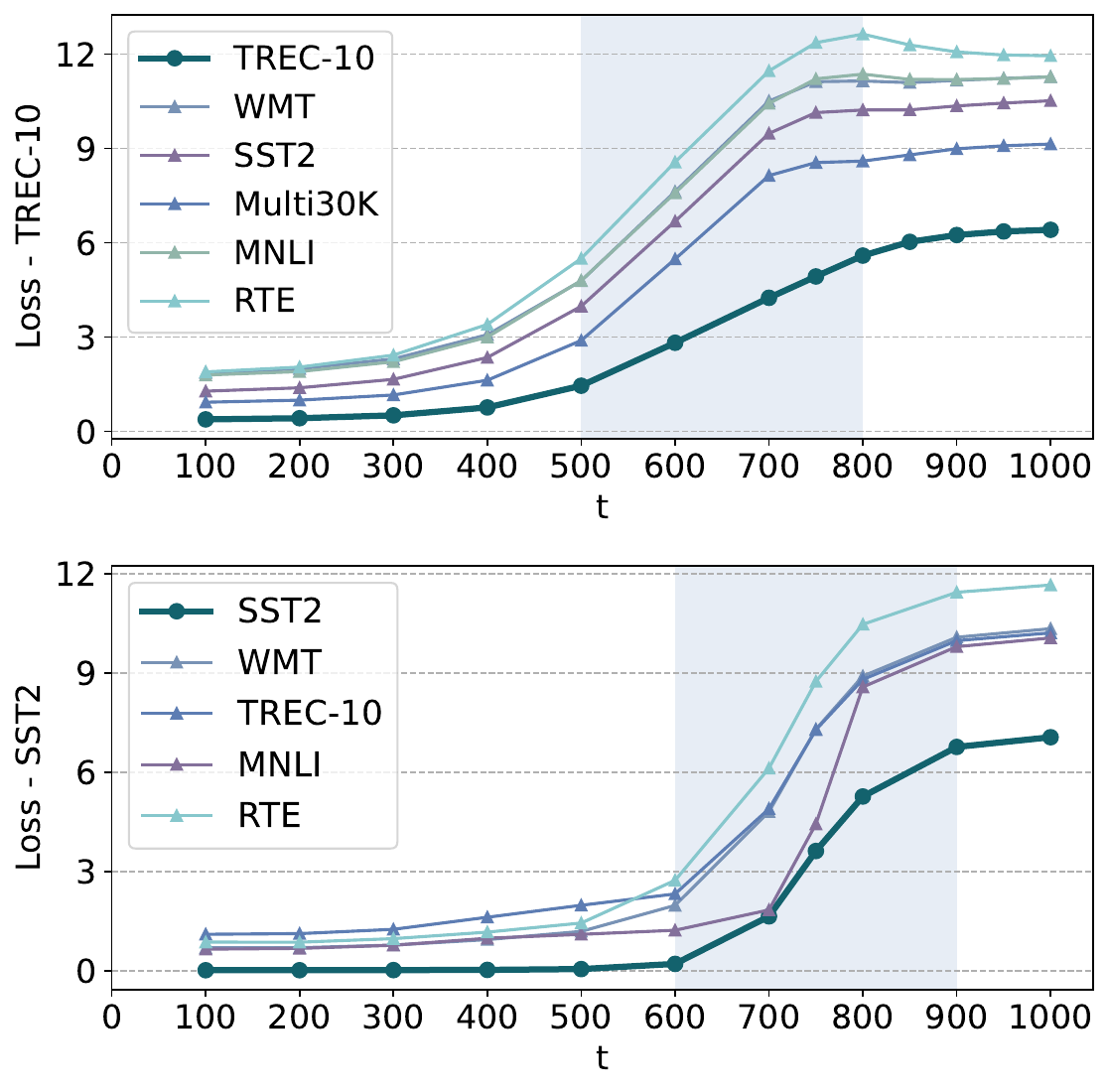}
\caption{The reconstruction loss on different OOD datasets by increasing diffusion step $t$.}
\label{t}
\end{figure}

As shown in Fig.~\ref{t}, the reconstruction loss of both ID and OOD data keep rising as $t$ increases.
More specifically, the loss of OOD data rises faster than ID data, indicating that OOD data is more sensitive to higher noise level.
The loss gap between ID/OOD is also increasing with $t$ and achieves the maximum when $t=700\sim800$, where model outputs start to look dissimilar from the inputs.
%we present examples of model reconstructions on OOD data when $t=500$ and $t=700$ respectively (Fig.\ref{t-case}).
%Outputs start to produce more dissimilar tokens when $t=700$.
For instance, only the high-frequency word ``what'' can be correctly restored in ID dataset when $t=800$ (Table~\ref{t-case}).
As for OOD data, even the common words like ``it'', ``just'' and ``as'' are wrongly restored.
The reason is that highly-noised inputs reserve very little information from the original sentences and the model begins to output unconditioned samples more than doing reconstructions~\cite{DBLP:journals/corr/abs-2211-07740}, which results in high reconstruction loss.
It shows that \model have worse OOD robustness when $t$ is large.
%But when $t$ is too large, the AUROC accuracy starts to fluctuate, indicating that much noise contributes little to OOD detection.
\begin{table}
\centering
\includegraphics[width=0.49\textwidth]{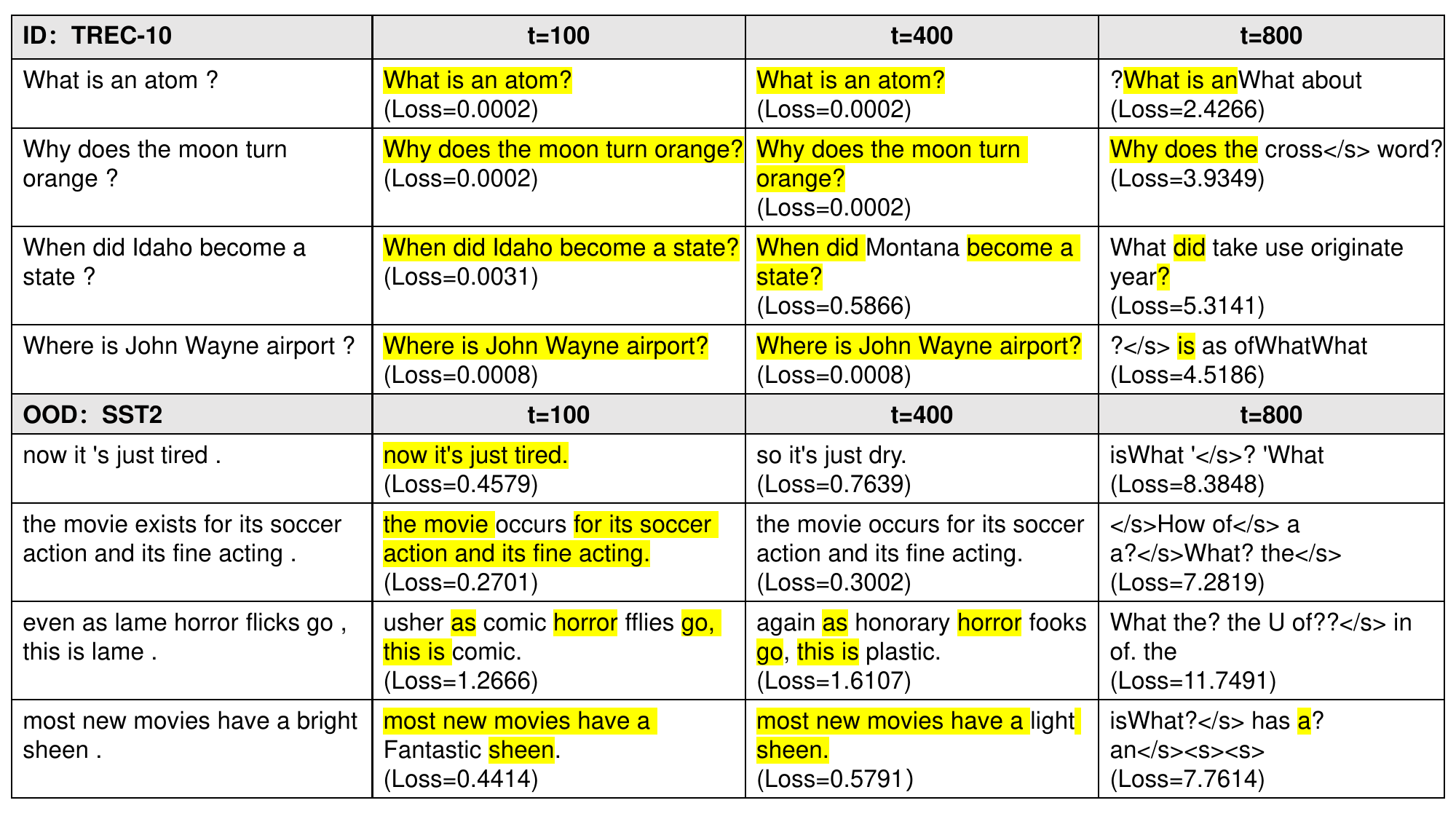}
\caption{Examples of the model outputs on ID (TREC-10) and OOD dataset (SST2) under different noise level $t$. The original sentences are in the leftmost column. Words painted yellow are correctly restored.}
\label{t-case}
\end{table}

\textbf{Models with more training steps are more resilience to higher noise level.}
The number of training steps $s$ is another important factor influencing the sensitivity of the model to noise.
We explore the model performance on ID/OOD data under different $t$ with the increase of $s$.

\begin{figure}
\centering
\includegraphics[width=0.48\textwidth]{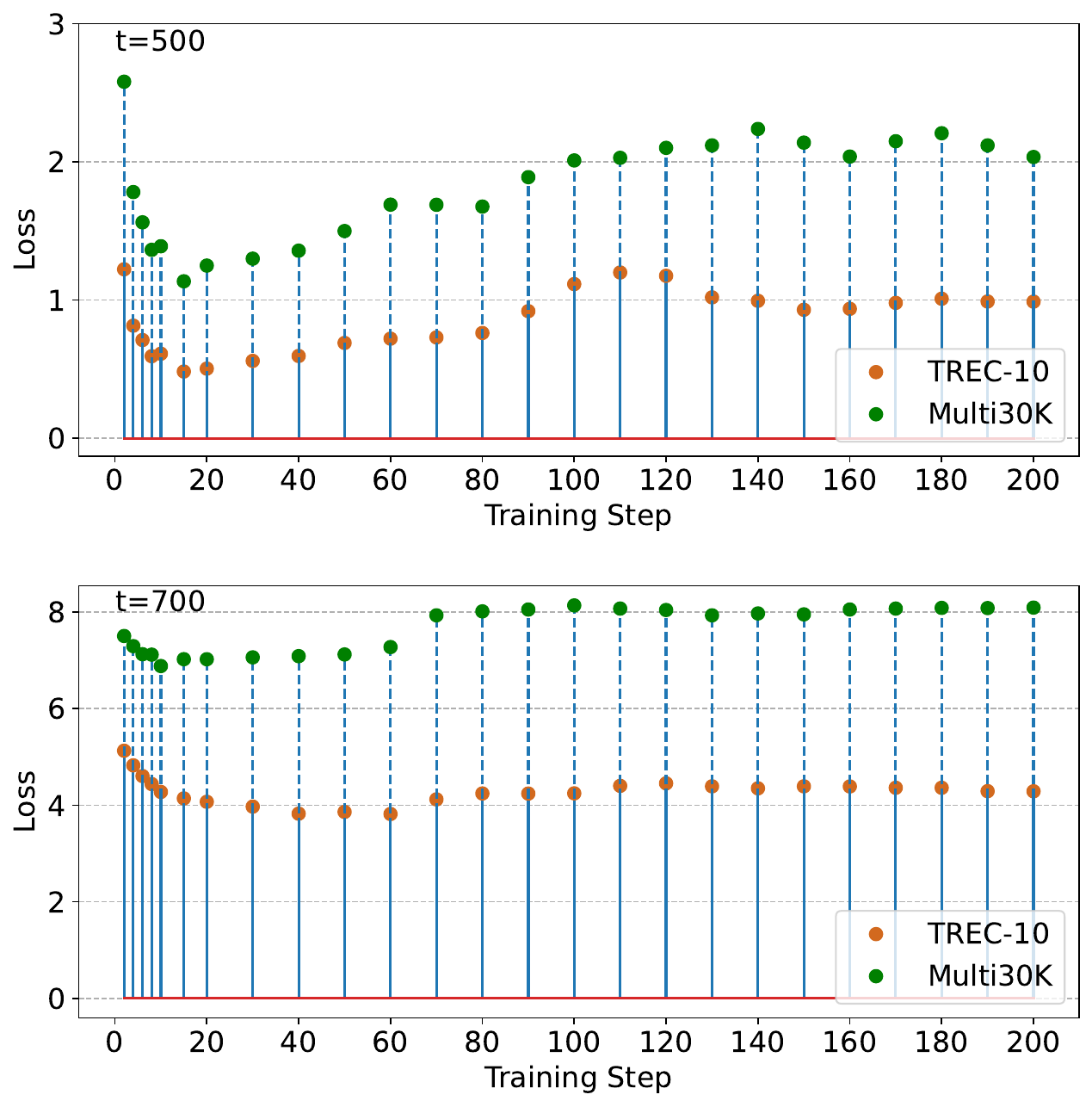}
\caption{Comparison of reconstruction loss for ID (TREC-10) and OOD (Multi30K) data by increasing training steps.}
\label{ckp}
\end{figure}

As noted in Fig.\ref{ckp}, when the added noise level is relative small ($t=500$), reconstruction loss of ID and OOD data reaches the minimum in small training steps ($s=20k\sim40k$) and then starts to rise and fluctuate.
When tested on higher noise level~($t=700$), the reconstruction loss of ID data basically declines and then remains stable.
However, reconstruction loss of OOD data goes up with the increase of $s$.
Such different behaviors verify that fully-trained models are more robust to higher noise level on ID data but the ability to restore OOD data is greatly degraded.

To verify the generation ability of fully-trained models, we test our models on SST2 dataset with different training steps and compare the score with MUCOLA-DISC~\cite{DBLP:conf/emnlp/KumarPT22}. The results are competitive with MUCOLA-DISC when reconstruction loss is relatively small ($s=160k$). This suggests that training models fully with reconstruction loss can guarantee good generation ability. 
\begin{table}
\centering
\footnotesize
\setlength{\tabcolsep}{1.5mm}
\begin{tabular}{c|cccccc}	%\hline
\hline
Method & Perplexity & Dist-1 & Dist-2 & Dist-3 & \makecell [c]{Reconstruc-\\tion loss}\\
\hline
GPT-2 & 38.60 & 0.64 & 0.90 & 0.88 & - \\
MUCOLA-DISC & 27.90 & 0.50 & 0.81 & 0.82 & -\\
Diffusion, s=20k & 10.66 & 0.42 & 0.56 & 0.69 & 3.5726\\
Diffusion, s=80k  & 30.21  &  0.62 & 0.78 & 0.81 & 1.8430 \\
Diffusion, s=160k & 33.76 & 0.63 & 0.80  & 0.83 & 1.4710\\
\hline
\end{tabular}
\caption{Scores of generation tasks on SST2 dataset compared with~\cite{DBLP:conf/emnlp/KumarPT22}.}
\label{data}
\setlength{\belowcaptionskip}{-0.2cm}	
\setlength{\abovecaptionskip}{-0.6cm}	
\end{table}

\textbf{Larger models are not good at reconstruction and have bad OOD robustness.}
It is proved that RoBERTa exhibits greater robustness than BERT on OOD data since it is pretrained on more diverse corpora~\cite{DBLP:conf/acl/HendrycksLWDKS20}.
To test this hypothesis, we also evaluate the reconstruction ability of diffusion models with more PLMs.

\begin{figure}
\centering
\setlength{\belowcaptionskip}{-0.25cm}
\includegraphics[width=0.48\textwidth]{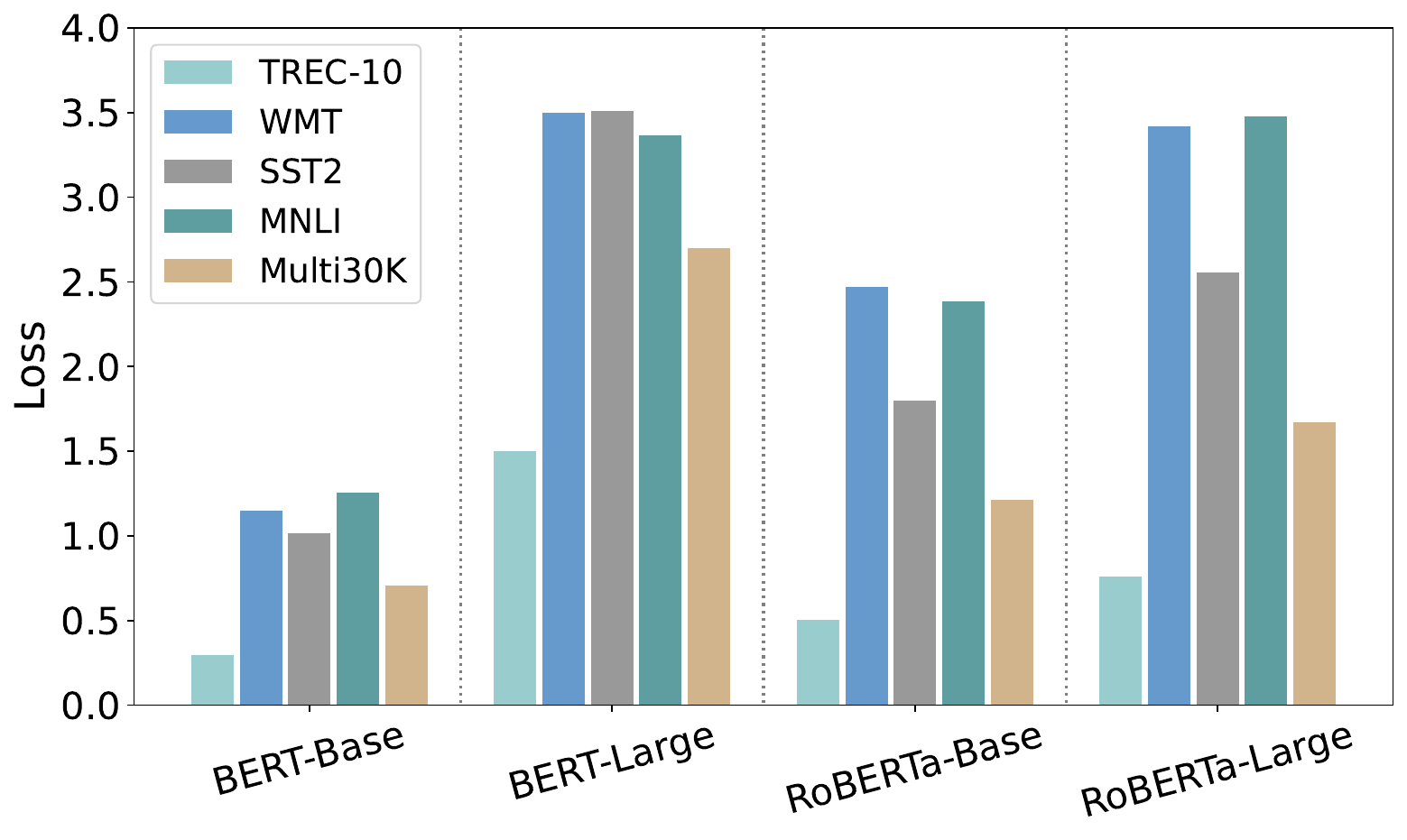}
\caption{Comparison of reconstruction loss for ID (TREC-10) and OOD datasets under different PLMs.}
\label{bert}
\end{figure}

As shown in Fig.\ref{bert}, the reconstruction loss of ID and OOD data when using RoBERTa is up to 2.8 times higher than that using BERT.
Besides, the loss of OOD data is up to 4.2 times higher than that of ID data when using BERT, while 4.7 times when using RoBERTa.
It shows that diffusion model with RoBERTa presents a greater gap of reconstruction loss between ID/OOD data and have worse OOD robustness than BERT. 
%It indicates that models pretrained on more diverse data do not reduce the ID/OOD reconstruction performance gap, but enlarge on the contrary~\cite{DBLP:conf/emnlp/Zhou0C21}.
%It indicates that diffusion models with RoBERTa have worse OOD robustness.

Besides, diffusion models with BERT/RoBERTa-Base have lower reconstruction loss than BERT/RoBERTa-Large.
The way in which the diffusion process adds noise determines this result.
Since the latent space dimension of the base model is lower than that of the large model, the influence of the added noise is smaller and the model can make better reconstructions conditioned on the remaining inputs.

\subsection{Impact of Data}

\textbf{Finetuning on more diverse data improves reconstruction quality, but not always.}
Since more diverse data improves OOD generalization of PLMs~\cite{DBLP:conf/acl/HendrycksLWDKS20}, we question if this assumption applies to diffusion as well.
To test it, we compare the OOD reconstruction performances when training with different ID datasets. 
We set a small $t=500$ in this experiment where the performance gap between different ID datasets is the most obvious.

As shown in Table~\ref{data}, compared to TREC-10, the model finetuned on SST2 dataset has a lower OOD reconstruction loss and a greater ability to restore the unfamiliar words that do not appear in ID dataset.
Since SST2 is a large dataset with more diverse data and covers 72.5\% OOD tokens, it suggests that finetuning on more diverse ID data improves OOD reconstruction performance~\cite{DBLP:conf/acl/JoshiH22}.

\begin{table}
\centering
\footnotesize
\setlength{\tabcolsep}{1.5mm}
\begin{tabular}{c|ccccc}	%\hline
\hline
Dataset & TREC-10 & SST2 & IMDB & 20NG\\
\hline
occur \& correct & 1524 & 2497 & 3464 & 3024 \\
occur \& wrong & 749 & 20 & 305 & 1169\\
not occur \& correct & 4 & 631 & 23 & 16\\
Avg loss  & 1.682  &  0.155 & 0.75 & 1.328 \\
%perplexity &   &  \\
\hline
vocab size & 9277 & 14409  & 41177 & 44944\\
overlap & 52.1\%  & 72.5\%  & 99.2\%  &  97.9\% \\
\hline
\end{tabular}
\caption{Statistics of the reconstruction results on OOD dataset (Multi30K).``not occur \& correct'' stands for the number of tokens that do not occur in ID dataset but are correctly restored. The vocabulary size of ID dataset and the number of overlapped OOD tokens are also reported.}
\label{data}
\setlength{\belowcaptionskip}{-0.2cm}	
\setlength{\abovecaptionskip}{-0.6cm}	
\end{table}

However, for the datasets IMDB and 20NG, although the value of ``overlap'' reaches 99.2\% and 97.9\% respectively (Table.~\ref{data}), models finetuned on these two datasets do not have better reconstruction performance.
By carefully investigating which aspect of data causes this anomalies, we compare the statistical characteristics of these data and observe that the average sentence length of IMDB and 20NG is 10$\sim$40 times larger than that of SST2 and TREC-10.
So we speculate that the OOD performance of diffusion model is also related to sentence length.

\begin{table*}
	\centerline{
	\includegraphics[width=15cm,height=8cm]{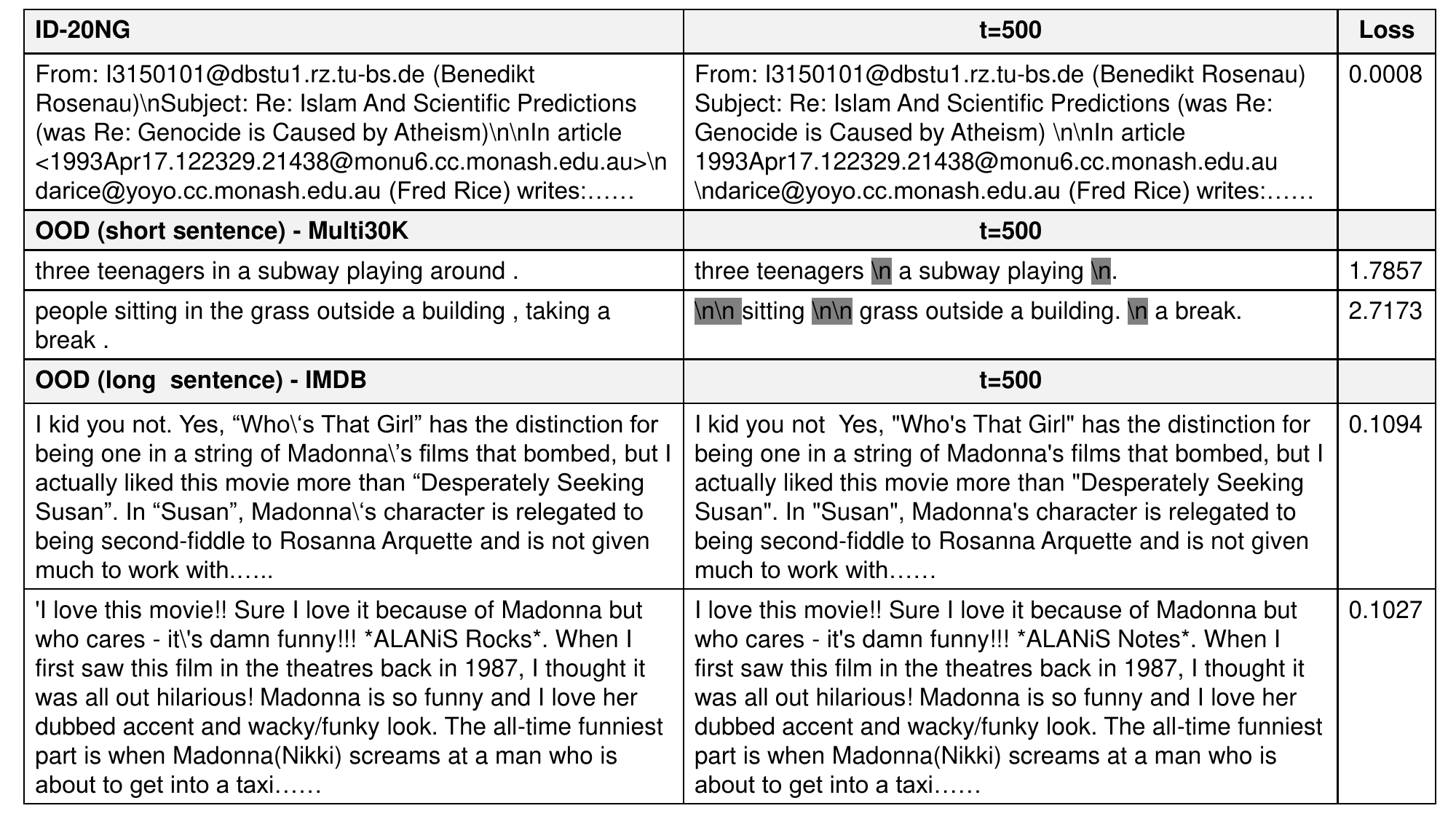}}
	\caption{Examples of model outputs on ID dataset (20NG) and OOD dataset with short sentences (Multi30K) and long sentences (IMDB). Wrongly-restored words are painted gray.}
	\label{app3}
\end{table*}

\textbf{Diffusion models with PLMs finetuned on long sentences are sensitive to length.}
To further test the hypothesis of length dependence, we explore the influence of sentence length on reconstruction quality.

\begin{figure}
\centering
\includegraphics[width=0.48\textwidth]{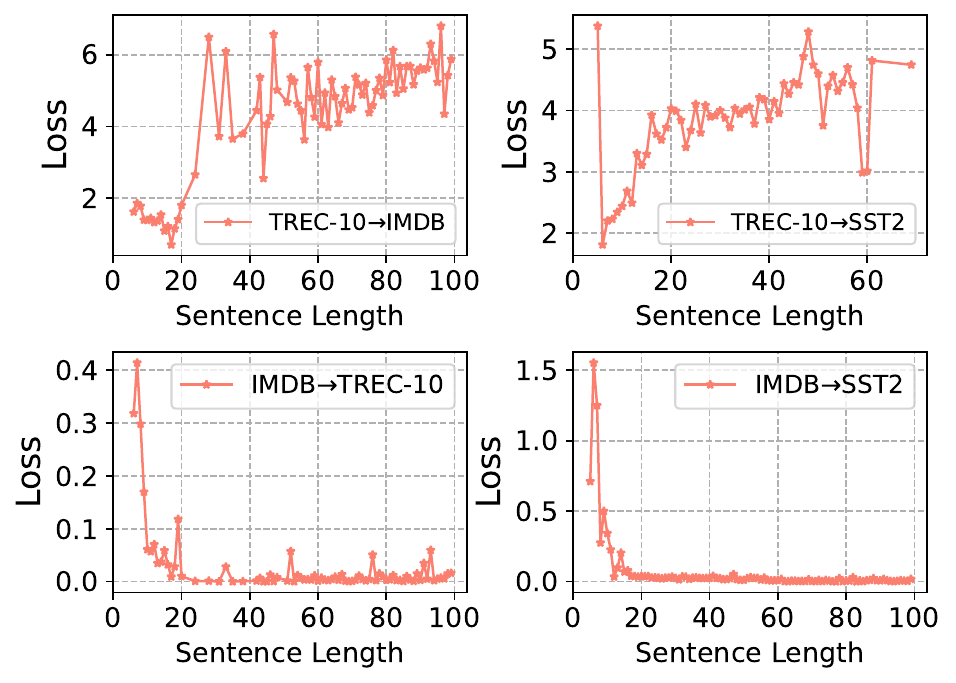}
\caption{Comparison of how the reconstruction loss of the model trained on long and short sentences vary with the increase of sentence length. ``TREC$\rightarrow{}$IMDB'' denotes that the model is finetuned on TREC-10 dataset and tested on IMDB dataset.}
\label{len}
\end{figure}

As shown in Fig.\ref{len}, when the model is finetuned on short sentences (TREC-10), the reconstruction loss of long sentence is overall higher than that of short sentence. The reason is that long sentences tend to contain more words not seen during training.
However, when the model is finetuned on long sentences (IMDB), the reconstruction loss of shorter sentence is at most 155 times higher than that of long sentence.
Even when tested on SST2 dataset which is not considered as OOD dataset from IMDB, the model fails to produce low-loss reconstructions on short sentences as well.
We call this phenomenon ``\emph{Length Bias}'', meaning that \model finetuned on long sentences are more sensitive to sentence length.

We give more examples to illustrate the \emph{``length bias''} problem using 20NG as ID dataset which has an average sentence length of 865.
%As shown in Table~\ref{app3}, most OOD tokens of both long and short sentences are correctly restored, further indicating that finetuned on more diverse data can make better reconstruction performance on OOD samples.
As shown in Table~\ref{app3}, sentence from ID dataset can be perfectly restored with an average loss of 0.0008.
It can also make near-perfect reconstructions on OOD dataset with long sentences~(IMDB).
However, when tested on short OOD sentences, even the common word ``in'' and ``the'' are not reconstructed correctly.
Besides, the loss of long OOD sentence is up to 27 times higher than that of short OOD sentences, indicating serious ``\emph{Length Bias}'' problem.
%(More details refer to \hyperref[a3]{A.3}.)

\begin{table*}[t]
\centering
\small
\begin{tabular}{c|ccccccc}
%\hline
\hline
\textbf{ID dataset} &  & &  & \textbf{OOD dataset} &\\
\hline
\small\verb|SST2| &\small\verb|IMDB| & \small\verb|TREC-10| & \small\verb|20NG|& \small\verb|MNLI|& \small\verb|RTE|&\small\verb|WMT16| & \small\verb|Multi30K| \\
Cosine$^{\ast}$ & - & 97.0/19.4 & 93.2/55.9 & 94.6/31.4 & 98.1/9.0 & 95.9/20.2 & 98.6/6.0\\
MLM & - & 96.0/32.1 & \textbf{100/0} & 99.3/0.1 & 99.8/0.1 & 98.2/12.9 & 68.9/76.2 \\
Hendrycks et al.~\cite{DBLP:conf/iclr/HendrycksG17}$^{\ast}$ & - & 91.8/61.3 & 93.6/52.1 & 84.6/68.4 & 89.2/59.7 & 84.0/69.4 & 90.2/57.3\\   % MSP
Liu et al.~\cite{DBLP:conf/nips/LiuWOL20}$^{\ast}$ & - & 91.5/63.1 & 93.4/52.5 & 83.6/68.7 & 87.4/62.4 & 82.4/70.6 &88.1/61.7\\  % Energy
Lee et al.~\cite{DBLP:conf/nips/LeeLLS18}$^{\ast}$ & - & 97.8/13.2 & 94.9/39.5 & 95.1/27.0 & 98.4/8.0 & 96.4/17.2 & 98.8/5.5 \\  % Maha
Zhou et al.~\cite{DBLP:conf/emnlp/Zhou0C21}$^{\ast}$ & - & 99.5/2.4 & \textbf{100/0} & \textbf{99.8}/0.4 & \textbf{100/0} & \textbf{99.9/0.1} & \textbf{100/0} \\
\hdashline
Diffusion & - & \textbf{100/0} & \textbf{100/0} & 99.6/1.5 & 99.9/0.4 & 99.4/2.2 & 79.5/70.4 \\
Diffusion+Maha & - & \textbf{100/0} & \textbf{100/0} & \textbf{99.8/0.5} & \textbf{100/0} & 99.7/1.1 & \textbf{100/0}\\
\hline
\hline
\small\verb|IMDB| &\small\verb|SST2| & \small\verb|TREC-10| & \small\verb|20NG|& \small\verb|MNLI|& \small\verb|RTE|&\small\verb|WMT16| & \small\verb|Multi30K| \\
Cosine$^{\ast}$ & - & 99.5/1.5 & 99.6/0.6 & 99.0/4.8 & 99.5/1.9 & 99.1/3.8 & 99.8/0.2\\
MLM & - & 99.7/3.1 & \textbf{100/0} & 87.8/36.1 & 90.0/33.8 & 86.8/36.4 & 99.7/0.1 \\
Hendrycks et al.~\cite{DBLP:conf/iclr/HendrycksG17}$^{\ast}$ & - & 94.9/37.4 & 96.0/28.4 & 93.1/51.4 & 93.9/49.9 & 93.4/50.9 & 96.4/25.7 \\
Liu et al.~\cite{DBLP:conf/nips/LiuWOL20}$^{\ast}$ & - & 94.0/57.7 & 95.6/32.1 & 92.4/55.4 & 93.3/54.7 & 92.7/57.6 & 95.6/34.3\\
Lee et al.~\cite{DBLP:conf/nips/LeeLLS18}$^{\ast}$ & - & \textbf{100/0} & 99.8/0.2 & 99.5/2.2 & 99.8/0.7 & 99.7/1.2 & 99.9/0\\
Zhou et al.~\cite{DBLP:conf/emnlp/Zhou0C21}$^{\ast}$ & - & \textbf{100/0} & \textbf{100/0} & \textbf{100/0.1}  & \textbf{100/0} & \textbf{100/0} & \textbf{100/0} \\
\hdashline
Diffusion & - & \textbf{100/0} & \textbf{100/0} & 99.4/2.7 & 98.3/8.9 & \textbf{100/0} & \textbf{100/0}\\
Diffusion+Maha & - & \textbf{100/0} & \textbf{100/0} & 99.2/0.5 & 99.4/3.3 & \textbf{100/0} & \textbf{100/0}\\
\hline
\hline
\small\verb|TREC-10| &\small\verb|SST2| & \small\verb|IMDB| & \small\verb|20NG|& \small\verb|MNLI|& \small\verb|RTE|&\small\verb|WMT16| & \small\verb|Multi30K| \\
Cosine$^{\ast}$ & 97.9/6.4 & 99.5/0.3 & 99.6/1.0 & 98.8/4.2 & 99.4/1.4 & 99.3/2.2 & 99.6/0.2\\
MLM  & 97.5/10.4 & \textbf{100/0} & \textbf{100/0} & 99.1/6.2 & 99.3/5.8 & 97.0/10.0 & 96.1/20.0\\
Hendrycks et al.~\cite{DBLP:conf/iclr/HendrycksG17}$^{\ast}$ & 97.1/14.5 & 98.9/2.9 & 98.2/7.2 & 97.0/13.6 & 98.6/5.2 & 97.9/8.5 & 99.1/1.3 \\
Liu et al.~\cite{DBLP:conf/nips/LiuWOL20}$^{\ast}$ & 94.8/28.5 & 98.9/4.5 & 99.0/6.6 & 97.3/15.5 & 98.8/5.5 & 98.2/10.2 & 99.2/1.8\\
Lee et al.~\cite{DBLP:conf/nips/LeeLLS18}$^{\ast}$ & 97.4/12.0 & 99.5/0.2 & 99.5/0.3 & 98.9/3.2 & 99.5/0.8 & 99.4/1.8 & \textbf{99.7}/0.3 \\
Zhou et al.~\cite{DBLP:conf/emnlp/Zhou0C21}$^{\ast}$ & 98.4/1.6 & 99.6/0 & 99.8/0 & 99.2/\textbf{0.7} & 99.6/\textbf{0.1} & 99.4/\textbf{0.5} & 99.5/\textbf{0} \\
\hdashline
Diffusion & 98.7/5.8 & \textbf{100/0} & \textbf{100/0} & 99.5/2.0 & 99.8/0.8 & 98.9/5.0 & 97.6/9.2\\
Diffusion+Maha & \textbf{99.7/0.8} & \textbf{100/0} & \textbf{100/0} & \textbf{99.7}/1.4 & \textbf{100/0.1} & \textbf{99.7}/2.4 & 99.6/1.2\\
\hline
\hline
\small\verb|20NG| &\small\verb|SST2| & \small\verb|IMDB| & \small\verb|TREC-10|& \small\verb|MNLI|& \small\verb|RTE|&\small\verb|WMT16| & \small\verb|Multi30K| \\
Cosine$^{\ast}$ & 99.7/0.9 & 98.6/6.4 & 98.7/8.3 & 97.5/13.9 & 95.2/21.9 & 96.8/16.4 & 98.3/7.0 \\
MLM  & 88.0/80.0 & 94.9/23.1 & 82.0/96.5 & 89.5/77.0 & 93.8/32.0 & 90.0/77.0 & 89.3/77.7\\
Hendrycks et al.~\cite{DBLP:conf/iclr/HendrycksG17}$^{\ast}$ & 98.6/9.0 & 95.9/25.3 & 95.1/35.0 & 94.1/36.0 & 90.3/49.0 & 92.7/40.5 & 95.7/18.9\\
Liu et al.~\cite{DBLP:conf/nips/LiuWOL20}$^{\ast}$ & 99.6/2.3 & 97.8/10.5 & 97.6/14.9 & 96.1/20.2 & 92.8/30.1 & 95.0/23.1 & 97.0/9.2 \\
Lee et al.~\cite{DBLP:conf/nips/LeeLLS18}$^{\ast}$ & 99.4/0.3 & 98.9/4.4 & 98.9/1.3 & 98.1/10.1 & 96.5/17.1 & 97.8/11.9 & 98.7/5.9 \\
Zhou et al.~\cite{DBLP:conf/emnlp/Zhou0C21}$^{\ast}$ & 99.5/1.2 & 99.0/4.7 & 99.6/1.4 & 98.4/9.6 & 98.2/11.1 & 98.5/7.5 & 99.1/6.9 \\
\hdashline
Diffusion & \textbf{100/0} & 99.9/0.4 & \textbf{100/0} & 99.9/0.1 & \textbf{100/0.0} & 99.8/0.1 & \textbf{100/0} \\
Diffusion+Maha & \textbf{100/0} & \textbf{100/0} & \textbf{100/0} & \textbf{100/0} & \textbf{100/0.0} & \textbf{100/0} &\textbf{100/0}\\
\hline
\end{tabular}
\caption{\label{acc}
The AUROC $\uparrow$ / FAR95 $\downarrow$ accuracy on different datasets when keeping $t=700$, $s=80K$ and $\lambda=0.99$. All data is averaged over ten times. $\ast$ denotes that results are taken from ~\cite{DBLP:conf/emnlp/Zhou0C21}.
}
\end{table*}

\section{Out-of-Distribution Detection}
\label{sec:detection}
In order to investigate to which extent diffusion influences PLMs on OOD data, in this section we present general performance of \model on OOD detection task.

\subsection{Baselines} 
We introduce the following methods as baselines of OOD detection.

$\bullet$ \textbf{Cosine Similarity} serves as a scoring function to measure the similarity of input representations. It is defined by the maximum cosine similarity of $\boldsymbol{h}$ to samples of the ID validation set $\boldsymbol{h}^{(dev)}$
\begin{equation}
\setlength{\abovedisplayskip}{3pt}
\setlength{\belowdisplayskip}{3pt}
d(x) = \mathop{\max}\limits_{i=1}^N cos(\boldsymbol{h}, \boldsymbol{h}_i^{(dev)})
\end{equation}

$\bullet$ \textbf{MLM} represents continuing training the pretrained transformers on ID data with Masked Language Modeling task~\cite{DBLP:conf/naacl/DevlinCLT19}.
Then we use the corresponding loss as a threshold to detect OOD data in a same way of diffusion-based OOD detection method.

$\bullet$ \textbf{Hendrycks et al.}~\cite{DBLP:conf/iclr/HendrycksG17} propose Maximum Softmax Probability (MSP) method by using the maximum class probability of a classifier with a softmax layer. The less confident classifier is, the higher the OOD score will be. It is defined by
\begin{equation}
\setlength{\abovedisplayskip}{3pt}
\setlength{\belowdisplayskip}{3pt}
d(x) = 1 - max_{j=1}^C \boldsymbol{p}_j,
%\label{maha}
\end{equation}
where $C$ is the number of classes.

$\bullet$ \textbf{Liu et al.}~\cite{DBLP:conf/nips/LiuWOL20} propose Energy Score method by using the softmax function as the ratio of the joint probability in $X \times Y$ to the probability in $X$, where $X$ and $Y$ denote the training corpus and its label set. A higher $d(x)$ means lower probability density in ID dataset, implying higher OOD likelihood. It calculates the following probability density
\begin{equation}
\setlength{\abovedisplayskip}{3pt}
\setlength{\belowdisplayskip}{3pt}
d(x) = -log \sum \limits_{j=1}^C exp(\boldsymbol{w}_j^T \boldsymbol{h}),
\end{equation}
where $\boldsymbol{w}_j$ is the weight of class $j$ in the softmax layer and $\boldsymbol{h}$ is the input of the softmax layer.

$\bullet$ \textbf{Lee et al.}~\cite{DBLP:conf/nips/LeeLLS18} propose to use Mahalanobis Distance~\cite{DBLP:conf/aaai/PodolskiyLBAP21} to determine the closeness of a sample to a set of samples belonging to the
class $c$ by modeling the ID features with class-conditional multivariate Gaussian distributions
\begin{equation}
\setlength{\abovedisplayskip}{3pt}
\setlength{\belowdisplayskip}{3pt}
d(x) = \mathop{\min}\limits_{c=1}^C(\boldsymbol{h(x)}-\boldsymbol{\mu}_c)^T\boldsymbol{\Sigma}^+(\boldsymbol{h(x)}-\boldsymbol{\mu}_c),
\label{maha}
\end{equation}
where $\boldsymbol{h(x)}$ is the vector representation of the sample $x$,
$\boldsymbol{\mu}_c$ is the mean vector of the training set of class $c$,
$\boldsymbol{\Sigma}$ is a shared covariance matrix and $\boldsymbol{\Sigma}^+$ is the pseudo-inverse of $\boldsymbol{\Sigma}$ defined by
\begin{equation}
\setlength{\abovedisplayskip}{3pt}
\setlength{\belowdisplayskip}{3pt}
\boldsymbol{\mu}_c = \frac{1}{N_c} \sum_{x\in D_{in}^c} \boldsymbol{h(x)},
%\label{maha}
\end{equation}
\begin{equation}
\setlength{\abovedisplayskip}{3pt}
\setlength{\belowdisplayskip}{3pt}
\boldsymbol{\Sigma} = \frac{1}{N} \sum_{c\in C}\sum_{x\in D_{in}^c} (\boldsymbol{h(x)}-\boldsymbol{\mu}_c)(\boldsymbol{h(x)}-\boldsymbol{\mu}_c)^T,
%\label{maha}
\end{equation}
where $N$ is the total number of the samples and $N_c$ is the number of samples
belonging to class $c$.

$\bullet$ \textbf{Zhou et al.}~\cite{DBLP:conf/emnlp/Zhou0C21} propose to finetune the Transformers with a contrastive loss.
%Instances from the same class have small L2 distances, forming compact clusters. %And the L2 distances of instances from different classes are larger than a margin.
By contrasting samples to those from different ID classes, it improves the compactness of representations. % where margin is defined as the maximum distance between pairs of instances from the same class.
The highest accuracy occurs when combining it with Mahalanobis Distance.

\subsection{Method}
We propose a diffusion-based detector which uses a score function $f(x) = L_{recon}(x)$ to measure the similarity of any sample $x$ with ID dataset by fixing a threshold $\gamma$. If $f(x)\le\gamma$, we classify $x$ as ID data. Otherwise, $x$ is regarded as OOD data.
Formally, we distinguish OOD data using the decision function:
\begin{equation}
%\begin{align*} 
\begin{split}
g(x,\gamma)= \left \{
\begin{array}{ll}
    0,     & f(x)\le\gamma\\
    1,     & f(x)\textgreater\gamma\\
\end{array}
\right.
\end{split}
%\end{align*}
\end{equation}
Since models learn to pay more attentions on token level after finetuned with diffusion, we further leverage Mahalanobis Distance method to add sentence level techniques. %, which is widely used in OOD detection to assess the closeness between a sample and a set of samples.
Following ~\cite{DBLP:conf/nips/LeeLLS18}, we define Mahalanobis Distance without using the class labels of the training set, i.e, treating the whole set as one class.
Eq.~\ref{maha} changes to
\begin{equation}
\setlength{\abovedisplayskip}{3pt}
\setlength{\belowdisplayskip}{3pt}
d(x) = (\boldsymbol{h(x)}-\boldsymbol{\mu}_s)^T\boldsymbol{\Sigma}^+(\boldsymbol{h(x)}-\boldsymbol{\mu}_s).
\label{maha2}
\end{equation}
\iffalse
where $h(x)$ is the vector representation of the sample $x$.
$\mu_s$ is the centroid of the training set.
$\Sigma$ is a shared covariance matrix and $\Sigma^+$ is the pseudo-inverse of $\Sigma$. 
defined as:
\begin{equation}
\setlength{\abovedisplayskip}{3pt}
\setlength{\belowdisplayskip}{3pt}
\mu_s = \frac{1}{N} \sum_{x\in D^{ID}} \boldsymbol{h(x)}
%\label{maha}
\end{equation}
\begin{equation}
\setlength{\abovedisplayskip}{3pt}
\setlength{\belowdisplayskip}{3pt}
\Sigma = \frac{1}{N} \sum_{x\in D^{ID}} (h(x)-\mu_s)(h(x)-\mu_s)^T
%\label{maha}
\end{equation}
\fi
Hence, we define our score function with a mixed of loss and Mahalanobis Distance score.
\begin{equation}
\setlength{\abovedisplayskip}{3pt}
\setlength{\belowdisplayskip}{3pt}
f(x) = \lambda \times L_{recon}(x) + (1 - \lambda) \times d(x)
%\label{maha}
\end{equation}
where $\lambda$ is a hyperparameter treated as a constant with regards to optimization.
%We fix a threshold $\gamma$ and classify $x$ as in domain if $s(x)\le\gamma$ or out of domain if $s(x)\textgreater\gamma$.

We use ``Diffusion'' to represent only using reconstruction loss as score function while ``Diffusion+Maha'' represents using both loss and Mahalanobis Distance.

\subsection{Metrics}
We apply two metrics that are commonly used in evaluating OOD detection performance~\cite{DBLP:conf/iclr/HendrycksG17} and both of them are threshold-independent.

$\bullet$ \textbf{AUROC}:
AUROC is the area under the Receiver Operating Characteristic curve.
%It is calculated as the true positive rate (TPR) against the false positive rate (FPR).
Higher AUROC value indicates better OOD detection performance. A random guessing detector model has an AUROC of 50\%.

$\bullet$ \textbf{FAR95}:
FAR95 is the probability that a negative example (OOD) to be mistakenly classified as positive (ID) when the true positive rate is 95\%. In this case, a lower value indicates better performance.

\subsection{Results And Analysis} 
\textbf{Overall.}
As shown in Table~\ref{acc}, the AUROC accuracies of ``Diffusion'' on most datasets are above 99\% and become SOTA.
It proves the effectiveness of \model on OOD detection task.
Specifically, the AUROC accuracy of ``Diffusion'' is up to 18\% higher than that of ``MLM'', indicating that \model degrade the generalization on OOD data.
One exception is that the accuracy when trained on SST2 and tested on Multi30K has a significant decline.
We suspect the reason is that SST2 contains more diverse data and has a similar sentence length with Multi30K.
So the reconstruction loss between ID and OOD data are too close to distinguish OOD data from ID data.
%The impact of $\lambda$ is discussed in \hyperref[a3]{A.4}.
%When adding Mahalanobis distance, most accuracies reach 100\% and become the SOTA.

\textbf{Impact of $\lambda$.} We test the impact of $\lambda$ on TREC-10 dataset while keeping $t=700$ and $s=80K$.
As shown in Table~\ref{lambda}, the accuracy reaches a peak at $\lambda=0.99$ and it declines by up to 2.27\% as the decrease of $\lambda$.
It indicates that the reconstruction loss plays a key role in OOD detection.
\begin{table}
\centering
\small
%\small
\begin{tabular}{c|cccc}	%\hline
\hline
Dataset & Multi30K & SST2 & WMT16 & RTE   \\
\hline
$\lambda=0.99$ & \textbf{98.32} & \textbf{98.57} & \textbf{99.33} & \textbf{99.53}\\
$\lambda=0.9$ & 97.57 & 96.83 & 98.73 & 98.40\\
$\lambda=0.7$ & 97.45 & 96.47 & 98.64 & 98.25\\
$\lambda=0.5$ & 97.43 & 96.40 & 98.63 & 98.22\\
$\lambda=0.3$ & 97.41 & 96.37 & 98.24 & 98.21\\
$\lambda=0.1$ & 97.41 & 96.30 & 98.24 & 98.20\\
\hline
\end{tabular}
\caption{
AUROC accuracy under different $\lambda$.}
\label{lambda}
\setlength{\belowcaptionskip}{-0.2cm}
\end{table}

\begin{table*}
\centering
\small
\begin{tabular}{c|ccccccc}
\hline
\small\verb|TREC-10| &\small\verb|SST2| & \small\verb|IMDB| & \small\verb|20NG|& \small\verb|MNLI|& \small\verb|RTE|&\small\verb|WMT16| & \small\verb|Multi30K| \\
\hline
Diffusion (T5-large)  & 98.9/4.6 & \textbf{100/0.0} & \textbf{100/0.0} & 99.6/1.4 & 99.9/0.4 & 99.3/2.6 & 98.2/6.6\\
Diffusion (Bart) & 99.3/3.2 & \textbf{100/0.0} & \textbf{100/0.0} & \textbf{100/0.0} & \textbf{100/0.0} & 99.8/0.6 & 99.4/1.8\\
\hline
\end{tabular}
\caption{\label{acc2}
The AUROC $\uparrow$ / FAR95 $\downarrow$ accuracy of T5-Large and BART finetuned with diffusion.
}
\end{table*}

\textbf{Impact of $\beta$.} We test the impact of $\beta$ on TREC-10 dataset while keeping $t=700$ and $s=80K$.
As shown in Table~\ref{beta}, the reconstruction loss of ID and OOD data decreases as $\beta$ gets smaller. But the gap between ID and OOD remains obvious. The model achieves the best results when $\beta$ increases linearly from $1e-4$ to $0.02$.

\begin{table}
\centering
\small
%\small
\begin{tabular}{cc|cccc}	%\hline
\hline
$\beta$ & & ID & Multi30K & RTE & WMT16   \\
\hline
\multirow{2}{*}{1e-3 to 0.2} & loss & 6.32 & 8.80 & 10.88 & 10.68 \\
\multirow{2}{*}{ } & AUROC & - &  95.3 & 99.4 & 98.4 \\
\hline
\multirow{2}{*}{1e-4 to 2e-2} & loss & 4.07 & 7.12 & 9.55 & 8.99 \\
\multirow{2}{*}{ } & AUROC & - & \textbf{96.9} & \textbf{99.7} & \textbf{98.7} \\
\hline
\multirow{2}{*}{1e-5 to 2e-3} & loss & 0.11 & 0.55 &  1.18 & 1.30 \\
\multirow{2}{*}{ } & AUROC & - & 90.6 & 97.2 & 95.8 \\
\hline
\end{tabular}
\caption{
Reconstruction loss and AUROC accuracy of ID (TREC-10) and OOD dataset under different scopes of $\beta$.}
\label{beta}
\setlength{\belowcaptionskip}{-0.2cm}
\end{table}

\begin{figure}
\centering
\setlength{\belowcaptionskip}{-0.25cm}
\includegraphics[width=7.7cm,height=4cm]{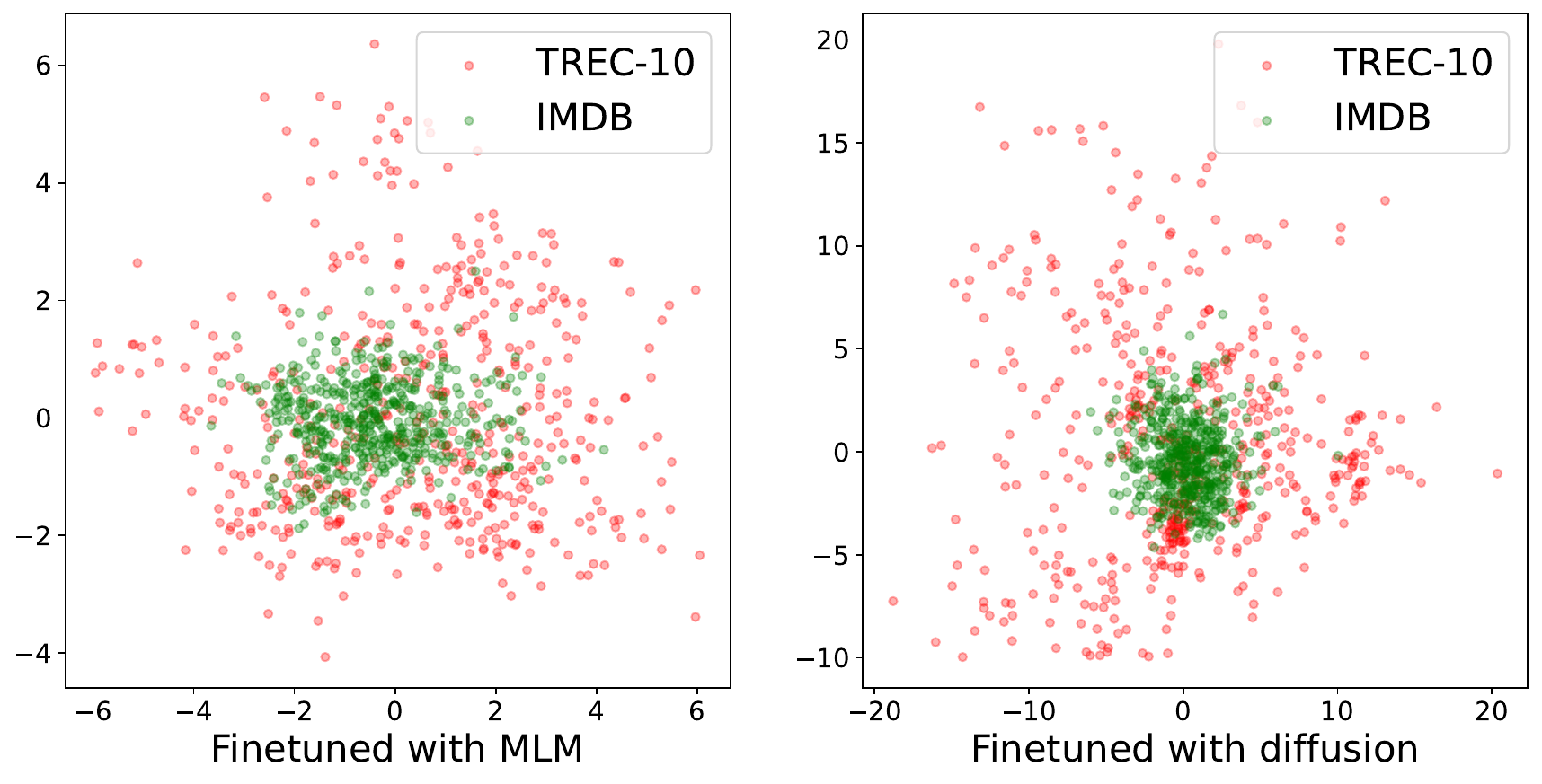}
\caption{2D visualization of average-pooled hidden-state sentence representations. Different colors represent different domains for each sentence.}
\label{scatter}
\end{figure}

\textbf{Visualization.}
By applying Gaussian Mixture Models (GMMs) to the learned representations, we draw the domain clusters in a 2D visualization as shown in Fig.\ref{scatter}.
Models are trained on TREC-10 as ID dataset.
After finetuning, ID data is scattered throughout the output space while OOD data tends to aggregate into a cluster.
Comparing with MLM, diffusion makes OOD data cluster in a more compact way like a Gauss.
ID data is more scattered since all sentences can be perfectly restored.
It further proves that \model are better OOD detectors.

\begin{figure}
\centering
\includegraphics[width=0.48\textwidth]{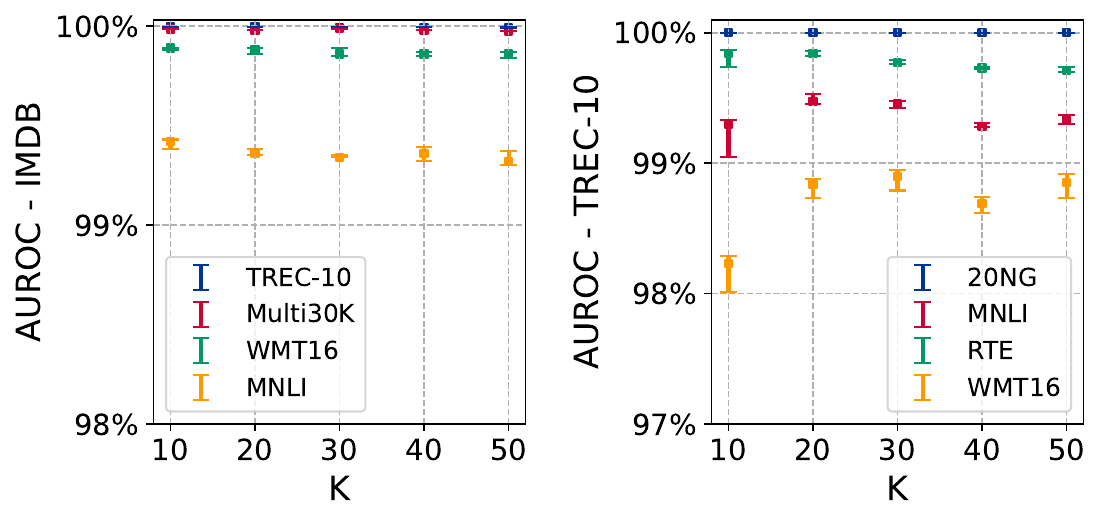}
\caption{Comparison of AUROC accuracy with errors under different K-shot scenarios.}
\label{fewshot}
\end{figure}

\textbf{Few-shot scenarios.}
In reality, many applications have very limited ID data. Hence, we test the robustness of diffusion-based detector finetuned on low-resource ID data.
%We test the robustness of diffusion-based detector in few-shot scenarios.
As shown in Fig.\ref{fewshot}, the AUROC accuracy remains high even when the number of training data is 10.
It indicates that a few number of ID data is enough for diffusion-based detector to detect OOD data.
%(More details refer to \hyperref[a4]{A.4}.)

\textbf{More PLMs.}
We further explore the OOD detection performance of T5-Large and BART finetuned with diffusion. TREC-10 is used as the ID dataset while keeping $s=80k$ and $t=800$. Results in Table~\ref{acc2} indicate that the OOD detection performances of T5-Large and BART outperform RoBERTa and become the new state-of-the-art, further illustrating that larger models finetuned with diffusion have worse OOD robustness.

We conclude that \model are effective on OOD detection, even in few-shot scenarios, though it shows a limitation when two datasets have high lexical coverage and similar sentence length. 
%On the other hand, it demonstrates good reconstruction ability on OOD data in such case.

\section{Related Work}

In real world scenarios, training and test examples are not always identical distributed, which makes models struggle to maintain the accuracy on out-of-distribution data. Hence, exploring whether the model is robust on OOD data is essential to build a safe and reliable system.

%prior works have revealed that ML models are vulnerable to adversarial examples. 
How to test OOD robustness is a complex and multi-faceted problem.
%We introduce several measurements that are widely-used on the study of OOD robustness.
Evaluating the accuracy on unseen in-distribution data~\cite{DBLP:conf/icml/MillerTRSKSLCS21} is a widely-used approach to study OOD robustness.
For instance, Angeli et al.~\cite{DBLP:journals/jbi/AngeliGDDWSDSWD22} explore the robustness of CNNs for text classification under distribution shifts. % and demonstrate that an ensemble of 12 CNNs can improve the OOD generalization.
It conducts two-phase learning by training and testing the model with different classes of ID data. 
Another line of work is evaluating OOD robustness by corrupting or perturbing inputs~\cite{DBLP:conf/iclr/HendrycksD19}. 
For instance, Wang et al.~\cite{DBLP:journals/corr/abs-2206-12361} explore the robustness of pretrained BNNs by fixing complex and nonlinear corruptions.
Moradi et al.~\cite{DBLP:conf/emnlp/MoradiS21} investigate the ability of language models in handling different types of character-level and word-level perturbations.
Zhang et al.~\cite{DBLP:conf/acl/0002PTK22} test how well the model learns to identify an unseen textual perturbation.

Different from prior works which conduct perturbation on discrete token space, we mainly focus on how diffusion in continuous space influences PLMs on OOD robustness. 
Since reconstructing a noised input is a self-supervised task and is a fundamental ability to measure diffusion model's performance, it is suitable for testing on data in various distributions, which serves naturally as a method to measure the model's OOD robustness. 
The reconstruction loss can not be directly compared with that of MLM since MLM can be seen as a one-step diffusion model on discrete state spaces~\cite{DBLP:conf/nips/AustinJHTB21} while we conduct diffusion on continuous latent space. Hence, we additionally use OOD detection task to further evaluate the OOD robustness~\cite{DBLP:journals/corr/abs-2211-07740} compared with other finetuning methods.

\section{Conclusion}
Understanding how diffusion influences pretrained language models on OOD data is crucial in NLP. In this work, we investigate the effect of \model on OOD robustness. The results indicate that diffusion reduces PLMs robustness, and provide a set of findings. Diffusion models with PLMs
(1) fail to make perfect reconstructions on partially-noised OOD examples;
(2) have serious length dependence when finetuned with long sentences;
(3) have worse reconstruction ability on OOD data when trained with larger models.
%(4) are robust in few-shot scenarios.
These findings have the following implications for other research:

\textit{Suggestions on applying diffusion.} Correctly adjusting the training process can alleviate robustness issues. Selecting smaller models, finetuning on more diverse data with a variety of sentence length is favorable for OOD robustness.

\textit{Diffusion is a new tool for OOD detection.} Diffusion models with PLMs achieve the state-of-the-art performance in most datasets, which is proved to be an effective tool. It provides a new idea to complete OOD sensitive work, such as outlier data detection, domain transfer, etc.

\section*{Ethical Statement}
In real-world scenarios, models may face heterogeneous samples which has severe semantic shifts from its training distributions.
The proposed work seeks to develop a comprehensive understanding of how diffusion influences PLMs on out-of-distribution data.
We believe that this study brings intellectual benefits to reliable application of diffusion models in the field of NLP. 
And it potentially has broader impacts to tasks of other areas.
There is not any direct societal consequence and all experiments are conducted on open datasets in this work.

\section*{Limitations}
As discussed in \S~\ref{sec:hyper}, the model size affects the robustness. Due to the constraints of computing power, it is difficult for us to incorporate some popular models with larger scales.
It should also be noted that our study is limited to English-only datasets. The pretrained language models we investigate are not in multilingual settings either. %While we have selected numerous and diverse tasks in this paper, it still represents a limited set. 
While results using automatic metrics give a fair idea of task performance, we would like to conduct a human evaluation in the near future.

%\clearpage


\ack This work was supported in part by the National Key R\&D Program of China 2020YFB1807800, in part by the National Natural Science Foundation of China under Grants (62201072, 62171057, 62071067), in part by the Ministry of Education and China Mobile Joint Fund (MCM20200202), Beijing University of Posts and Telecommunications-China Mobile Research Institute Joint Innovation Center.

\bibliography{diffusion}

\begin{thebibliography}{10}

\bibitem{DBLP:journals/jbi/AngeliGDDWSDSWD22}
Kevin~De Angeli, Shang Gao, Ioana Danciu, Eric~B. Durbin, Xiao{-}Cheng Wu,
  Antoinette Stroup, Jennifer~A. Doherty, Stephen~M. Schwartz, Charles Wiggins,
  Mark Damesyn, Linda Coyle, Lynne Penberthy, Georgia~D. Tourassi, and
  Hong{-}Jun Yoon, `Class imbalance in out-of-distribution datasets: Improving
  the robustness of the textcnn for the classification of rare cancer types',
  {\em J. Biomed. Informatics}, {\bf 125},  103957, (2022).

\bibitem{DBLP:conf/nips/AustinJHTB21}
Jacob Austin, Daniel~D. Johnson, Jonathan Ho, Daniel Tarlow, and Rianne van~den
  Berg, `Structured denoising diffusion models in discrete state-spaces', in
  {\em NeurIPS 2021}, pp. 17981--17993, (2021).

\bibitem{DBLP:conf/wmt/BojarCFGHHJKLMN16}
Ondrej Bojar, Rajen Chatterjee, Christian Federmann, Yvette Graham, Barry
  Haddow, Matthias Huck, Antonio Jimeno{-}Yepes, Philipp Koehn, Varvara
  Logacheva, Christof Monz, Matteo Negri, Aur{\'{e}}lie N{\'{e}}v{\'{e}}ol,
  Mariana~L. Neves, Martin Popel, Matt Post, Raphael Rubino, Carolina Scarton,
  Lucia Specia, Marco Turchi, Karin Verspoor, and Marcos Zampieri, `Findings of
  the 2016 conference on machine translation', in {\em Proceedings of the First
  Conference on Machine Translation, {WMT} 2016, colocated with {ACL} 2016,
  August 11-12, Berlin, Germany}, pp. 131--198, (2016).

\bibitem{DBLP:conf/mlcw/DaganGM05}
Ido Dagan, Oren Glickman, and Bernardo Magnini, `The {PASCAL} recognising
  textual entailment challenge', in {\em Machine Learning Challenges,
  Evaluating Predictive Uncertainty, Visual Object Classification and
  Recognizing Textual Entailment, First {PASCAL} Machine Learning Challenges
  Workshop, {MLCW} 2005, Southampton, UK, April 11-13, 2005, Revised Selected
  Papers}, volume 3944 of {\em Lecture Notes in Computer Science}, pp.
  177--190, (2005).

\bibitem{DBLP:conf/naacl/DevlinCLT19}
Jacob Devlin, Ming{-}Wei Chang, Kenton Lee, and Kristina Toutanova, `{BERT:}
  pre-training of deep bidirectional transformers for language understanding',
  in {\em {NAACL-HLT} 2019, Volume 1 (Long and Short Papers)}, pp. 4171--4186,
  (2019).

\bibitem{elliott2016multi30k}
Desmond Elliott, Stella Frank, Khalil Sima'an, and Lucia Specia, `Multi30k:
  Multilingual english-german image descriptions', {\em arXiv preprint
  arXiv:1605.00459}, (2016).

\bibitem{DBLP:journals/corr/abs-2210-08933}
Shansan Gong, Mukai Li, Jiangtao Feng, Zhiyong Wu, and Lingpeng Kong,
  `Diffuseq: Sequence to sequence text generation with diffusion models', {\em
  CoRR}, {\bf abs/2210.08933}, (2022).

\bibitem{DBLP:journals/corr/abs-2211-07740}
Mark~S. Graham, Walter H.~L. Pinaya, Petru{-}Daniel Tudosiu, Parashkev Nachev,
  S{\'{e}}bastien Ourselin, and M.~Jorge Cardoso, `Denoising diffusion models
  for out-of-distribution detection', {\em CoRR}, {\bf abs/2211.07740}, (2022).

\bibitem{DBLP:journals/corr/abs-2211-15029}
Zhengfu He, Tianxiang Sun, Kuanning Wang, Xuanjing Huang, and Xipeng Qiu,
  `Diffusionbert: Improving generative masked language models with diffusion
  models', {\em CoRR}, {\bf abs/2211.15029}, (2022).

\bibitem{DBLP:conf/iclr/HendrycksD19}
Dan Hendrycks and Thomas~G. Dietterich, `Benchmarking neural network robustness
  to common corruptions and perturbations', in {\em {ICLR} 2019, New Orleans,
  LA, USA, May 6-9, 2019}. OpenReview.net, (2019).

\bibitem{DBLP:conf/iclr/HendrycksG17}
Dan Hendrycks and Kevin Gimpel, `A baseline for detecting misclassified and
  out-of-distribution examples in neural networks', in {\em {ICLR} 2017,
  Toulon, France, April 24-26, 2017, Conference Track Proceedings}, (2017).

\bibitem{DBLP:conf/acl/HendrycksLWDKS20}
Dan Hendrycks, Xiaoyuan Liu, Eric Wallace, Adam Dziedzic, Rishabh Krishnan, and
  Dawn Song, `Pretrained transformers improve out-of-distribution robustness',
  in {\em {ACL} 2020, Online, July 5-10, 2020}, pp. 2744--2751, (2020).

\bibitem{DBLP:conf/nips/HoJA20}
Jonathan Ho, Ajay Jain, and Pieter Abbeel, `Denoising diffusion probabilistic
  models', in {\em NeurIPS 2020, December 6-12, 2020, virtual}, (2020).

\bibitem{DBLP:conf/acl/JoshiH22}
Nitish Joshi and He~He, `An investigation of the (in)effectiveness of
  counterfactually augmented data', in {\em {ACL} 2022}, pp. 3668--3681,
  (2022).

\bibitem{DBLP:journals/corr/KingmaB14}
Diederik~P. Kingma and Jimmy Ba, `Adam: {A} method for stochastic
  optimization', in {\em {ICLR} 2015, San Diego, CA, USA, May 7-9, 2015,
  Conference Track Proceedings}, (2015).

\bibitem{DBLP:conf/emnlp/KumarPT22}
Sachin Kumar, Biswajit Paria, and Yulia Tsvetkov, `Gradient-based constrained
  sampling from language models', in {\em {EMNLP} 2022}, pp. 2251--2277,
  (2022).

\bibitem{DBLP:conf/icml/Lang95}
Ken Lang, `Newsweeder: Learning to filter netnews', in {\em Machine Learning,
  Proceedings of the Twelfth International Conference on Machine Learning,
  Tahoe City, California, USA, July 9-12, 1995}, pp. 331--339, (1995).

\bibitem{DBLP:conf/nips/LeeLLS18}
Kimin Lee, Kibok Lee, Honglak Lee, and Jinwoo Shin, `A simple unified framework
  for detecting out-of-distribution samples and adversarial attacks', in {\em
  NeurIPS 2018, December 3-8, 2018, Montr{\'{e}}al, Canada}, pp. 7167--7177,
  (2018).

\bibitem{li2022diffusionlm}
Xiang~Lisa Li, John Thickstun, Ishaan Gulrajani, Percy Liang, and Tatsunori
  Hashimoto, `Diffusion-{LM} improves controllable text generation', in {\em
  NeurIPS 2022, November 28 - December 9, 2022, New Orleans}, (2022).

\bibitem{DBLP:conf/coling/LiR02}
Xin Li and Dan Roth, `Learning question classifiers', in {\em {COLING} 2002,
  Howard International House and Academia Sinica, Taipei, Taiwan, August 24 -
  September 1, 2002}, (2002).

\bibitem{DBLP:conf/nips/LiuWOL20}
Weitang Liu, Xiaoyun Wang, John~D. Owens, and Yixuan Li, `Energy-based
  out-of-distribution detection', in {\em NeurIPS 2020, December 6-12, 2020,
  virtual}, (2020).

\bibitem{DBLP:conf/acl/MaasDPHNP11}
Andrew~L. Maas, Raymond~E. Daly, Peter~T. Pham, Dan Huang, Andrew~Y. Ng, and
  Christopher Potts, `Learning word vectors for sentiment analysis', in {\em
  The 49th Annual Meeting of the Association for Computational Linguistics:
  Human Language Technologies, Proceedings of the Conference, 19-24 June, 2011,
  Portland, Oregon, {USA}}, pp. 142--150, (2011).

\bibitem{DBLP:conf/icml/MillerTRSKSLCS21}
John Miller, Rohan Taori, Aditi Raghunathan, Shiori Sagawa, Pang~Wei Koh,
  Vaishaal Shankar, Percy Liang, Yair Carmon, and Ludwig Schmidt, `Accuracy on
  the line: on the strong correlation between out-of-distribution and
  in-distribution generalization', in {\em Proceedings of the 38th
  International Conference on Machine Learning, {ICML} 2021, 18-24 July 2021,
  Virtual Event}, volume 139 of {\em Proceedings of Machine Learning Research},
  pp. 7721--7735. {PMLR}, (2021).

\bibitem{DBLP:conf/emnlp/MoradiS21}
Milad Moradi and Matthias Samwald, `Evaluating the robustness of neural
  language models to input perturbations', in {\em {EMNLP} 2021, Virtual Event
  / Punta Cana, Dominican Republic, 7-11 November, 2021}, pp. 1558--1570,
  (2021).

\bibitem{DBLP:conf/icml/NicholDRSMMSC22}
Alexander~Quinn Nichol, Prafulla Dhariwal, Aditya Ramesh, Pranav Shyam, Pamela
  Mishkin, Bob McGrew, Ilya Sutskever, and Mark Chen, `{GLIDE:} towards
  photorealistic image generation and editing with text-guided diffusion
  models', in {\em {ICML} 2022}, volume 162, pp. 16784--16804, (2022).

\bibitem{DBLP:conf/aaai/PodolskiyLBAP21}
Alexander Podolskiy, Dmitry Lipin, Andrey Bout, Ekaterina Artemova, and Irina
  Piontkovskaya, `Revisiting mahalanobis distance for transformer-based
  out-of-domain detection', in {\em {AAAI} 2021, {IAAI} 2021, {EAAI} 2021,
  Virtual Event, February 2-9, 2021}, pp. 13675--13682, (2021).

\bibitem{DBLP:journals/corr/abs-2204-06125}
Aditya Ramesh, Prafulla Dhariwal, Alex Nichol, Casey Chu, and Mark Chen,
  `Hierarchical text-conditional image generation with {CLIP} latents', {\em
  CoRR}, {\bf abs/2204.06125}, (2022).

\bibitem{DBLP:conf/emnlp/SocherPWCMNP13}
Richard Socher, Alex Perelygin, Jean Wu, Jason Chuang, Christopher~D. Manning,
  Andrew~Y. Ng, and Christopher Potts, `Recursive deep models for semantic
  compositionality over a sentiment treebank', in {\em {EMNLP} 2013, {A}
  meeting of SIGDAT, a Special Interest Group of the {ACL}}, pp. 1631--1642,
  (2013).

\bibitem{DBLP:journals/corr/abs-2211-04236}
Robin Strudel, Corentin Tallec, Florent Altch{\'{e}}, Yilun Du, Yaroslav Ganin,
  Arthur Mensch, Will Grathwohl, Nikolay Savinov, Sander Dieleman, Laurent
  Sifre, and R{\'{e}}mi Leblond, `Self-conditioned embedding diffusion for text
  generation', {\em CoRR}, {\bf abs/2211.04236}, (2022).

\bibitem{DBLP:journals/corr/abs-2206-12361}
Xi~Wang and Laurence Aitchison, `Out of distribution robustness with
  pre-trained bayesian neural networks', {\em CoRR}, {\bf abs/2206.12361},
  (2022).

\bibitem{DBLP:conf/naacl/WilliamsNB18}
Adina Williams, Nikita Nangia, and Samuel~R. Bowman, `A broad-coverage
  challenge corpus for sentence understanding through inference', in {\em
  {NAACL-HLT} 2018, Volume 1 (Long Papers)}, pp. 1112--1122, (2018).

\bibitem{DBLP:conf/emnlp/WolfDSCDMCRLFDS20}
Thomas Wolf, Lysandre Debut, Victor Sanh, Julien Chaumond, Clement Delangue,
  Anthony Moi, Pierric Cistac, Tim Rault, R{\'{e}}mi Louf, Morgan Funtowicz,
  Joe Davison, Sam Shleifer, Patrick von Platen, Clara Ma, Yacine Jernite,
  Julien Plu, Canwen Xu, Teven~Le Scao, Sylvain Gugger, Mariama Drame, Quentin
  Lhoest, and Alexander~M. Rush, `Transformers: State-of-the-art natural
  language processing', in {\em {EMNLP} 2020 - Demos, Online, November 16-20,
  2020}, pp. 38--45, (2020).

\bibitem{DBLP:conf/acl/0002PTK22}
Yunxiang Zhang, Liangming Pan, Samson Tan, and Min{-}Yen Kan, `Interpreting the
  robustness of neural {NLP} models to textual perturbations', in {\em Findings
  of the Association for Computational Linguistics: {ACL} 2022, Dublin,
  Ireland, May 22-27, 2022}, pp. 3993--4007, (2022).

\bibitem{DBLP:conf/emnlp/Zhou0C21}
Wenxuan Zhou, Fangyu Liu, and Muhao Chen, `Contrastive out-of-distribution
  detection for pretrained transformers', in {\em {EMNLP} 2021}, pp.
  1100--1111, (2021).

\end{thebibliography}
\end{document}